\documentclass[twoside,11pt]{article}

\usepackage[preprint]{jmlr2e}

\usepackage{booktabs}
\usepackage{amsmath}

\def\code#1{\texttt{#1}}

\usepackage{listings}
\usepackage{xcolor}
\definecolor{comment_color}{RGB}{0,128,0}
\definecolor{number_color}{RGB}{35,120,147}
\definecolor{string_color}{RGB}{163,21,21}
\definecolor{background_color}{RGB}{255,255,255}
\definecolor{keyword_color}{RGB}{0,0,255}

\lstdefinestyle{mystyle}{
	backgroundcolor=\color{background_color},
	commentstyle=\color{comment_color},
	keywordstyle=\color{keyword_color},
	numberstyle=\tiny\color{number_color},
	stringstyle=\color{string_color},
	basicstyle=\ttfamily\footnotesize,
	breakatwhitespace=false,
	breaklines=true,
	captionpos=b,
	keepspaces=true,
	numbers=left,
	numbersep=5pt,
	showspaces=false,
	showstringspaces=false,
	showtabs=false,
	tabsize=2
}
\usepackage{lastpage}
\jmlrheading{23}{2022}{1-\pageref{LastPage}}{12/21; Revised
11/22}{12/22}{21-1518}{F\'{a}bio M. Miranda, Niklas K{\"o}hnecke and Bernhard Y. Renard}

\ShortHeadings{HiClass: a Python Library for Local Hierarchical Classification}{Miranda, K{\"o}hnecke and Renard}
\firstpageno{1}

\begin{document}

\title{HiClass: a Python Library for Local Hierarchical Classification Compatible with Scikit-learn}

\author{\name{F\'{a}bio M. Miranda\thanks{These authors contributed equally to this work.} \addtocounter{footnote}{-1}\addtocounter{Hfootnote}{-1}} \email fabio.malchermiranda@hpi.de \\
       \addr Hasso Plattner Institute, 
       Digital Engineering Faculty,
       University of Potsdam,\\ 14482 Potsdam, Germany \\
       \addr Department of Mathematics and Computer Science, Free University of Berlin,\\ 14195 Berlin, Germany
       \AND
       \name{Niklas K{\"o}hnecke\footnotemark} \email niklas.koehnecke@student.hpi.uni-potsdam.de \\
       \addr Hasso Plattner Institute, 
       Digital Engineering Faculty, University of Potsdam, \\14482 Potsdam, Germany
       \AND
       \name Bernhard Y. Renard \email bernhard.renard@hpi.de \\
       \addr Hasso Plattner Institute, 
       Digital Engineering Faculty,
       University of Potsdam,\\ 14482 Potsdam, Germany}

\editor{Alexandre Gramfort}

\maketitle

\begin{abstract}
\texttt{HiClass} is an open-source Python library for local hierarchical classification entirely compatible with \texttt{scikit-learn}. It contains implementations of the most common design patterns for hierarchical machine learning models found in the literature, that is, the local classifiers per node, per parent node and per level. Additionally, the package contains implementations of hierarchical metrics, which are more appropriate for evaluating classification performance on hierarchical data. The documentation includes installation and usage instructions, examples within tutorials and interactive notebooks, and a complete description of the API. \texttt{HiClass} is released under the simplified BSD license, encouraging its use in both academic and commercial environments. Source code and documentation are available at \url{https://github.com/scikit-learn-contrib/hiclass}.
\end{abstract}


\section{Introduction}

\begin{figure}[!htb]
\centerline{\includegraphics[width=\linewidth]{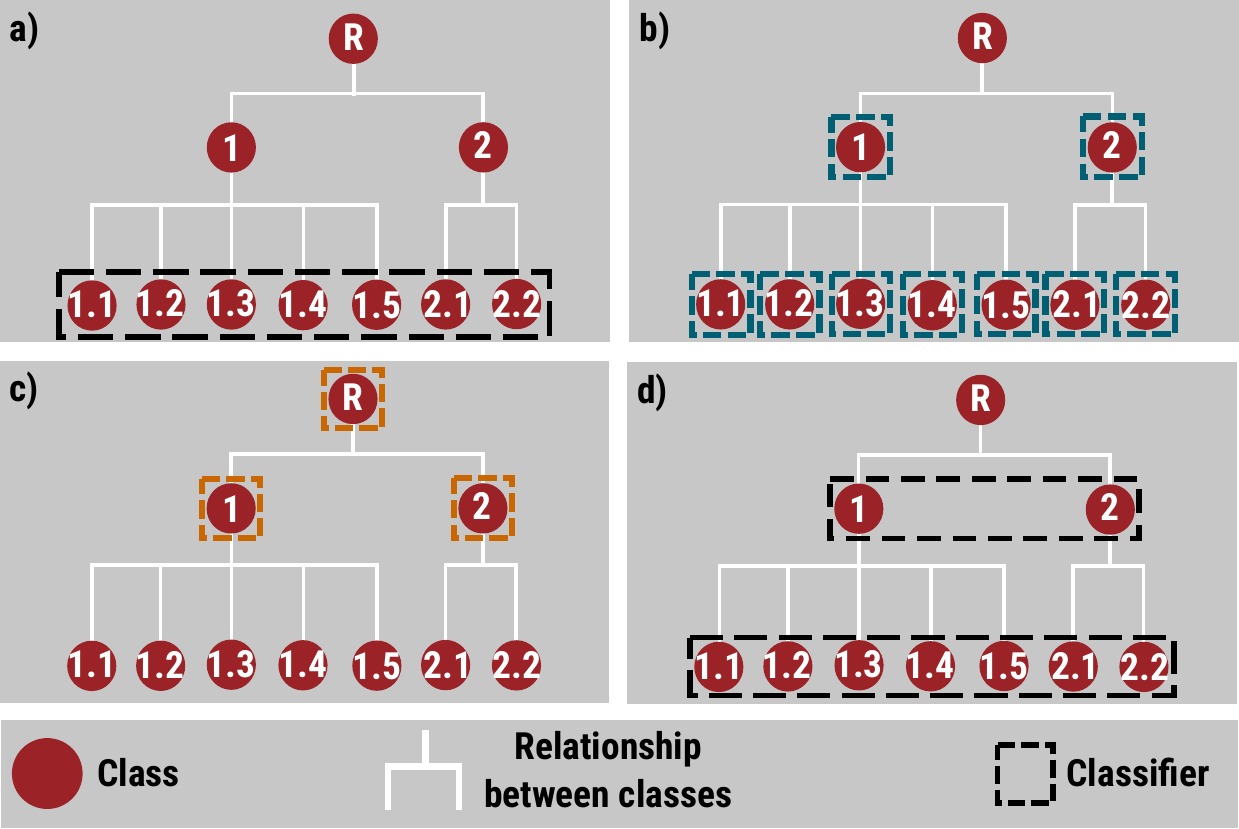}}
\caption{Depiction of the flat approach and the local hierarchical classification models implemented in \texttt{HiClass}, for a two-level hierarchy. a) Flat approach, where a single multi-class classifier is trained to predict only the leaf nodes in hierarchical data, completely ignoring previous levels of the hierarchy. b) Local classifier per node, where several binary classifiers are trained for each existing node in the hierarchy, excluding the root node. c) Local classifier per parent node, where various multi-class classifiers are trained for each parent node in the hierarchy, with the goal of predicting the children nodes. d) Local classifier per level, where a multi-class classifier is trained for each level in the hierarchy.}\label{fig:classifiers}
\end{figure}

Many classification problems across different application domains can be naturally modeled hierarchically (\autoref{fig:classifiers} and appendix Figures \ref{fig:music}-\ref{fig:phylogeny}), typically in the form of trees or directed acyclic graphs \citep{silla2011survey}. Examples of hierarchical classification problems are vast, ranging from musical genre classification \citep{ariyaratne2012novel, iloga2018sequential} to text categorization \citep{javed2021hierarchical, ma2022hybrid}, taxonomic classification of viral sequences in metagenomic data \citep{shang2021cheer} and identification of COVID-19 in chest X-ray images \citep{pereira2020covid}.

Nonetheless, many classifiers proposed in the literature are designed to completely ignore the existing hierarchy between classes by usually predicting only leaf nodes in a methodology known as flat approach. Although easy to implement, the flat approach is incapable of dealing with problems where making a prediction for leaf nodes is not mandatory. Furthermore, since they consider the hierarchy during training, hierarchical models generally produce better results when compared with flat approaches, consistently achieving improvements in predictive performance \citep{silla2011survey}. Hence, in this manuscript we introduce \texttt{HiClass}, a Python library that implements the most common patterns for local hierarchical classifiers (see \autoref{fig:classifiers} and Appendix C for more details), which can be employed in different application domains where the data can be hierarchically structured in the shape of trees or directed acyclic graphs. While the full hierarchical structure is ideally known, classification is also possible with only partially known structures and missing values in the outermost levels of the hierarchy.

\section{Overview and Design}

The API of \texttt{HiClass} is designed to be compatible with the \texttt{scikit-learn} API \citep{pedregosa2011scikit}, offering users a familiar API to train hierarchical models via \textit{fit()}, store trained models with \texttt{pickle}, predict labels with the \textit{predict()} method, and create machine learning pipelines. Moreover, each local classifier in \texttt{HiClass} is, by default, \texttt{scikit-learn}'s \textit{BaseEstimator} class, enabling users to employ any of the classifiers already implemented in \texttt{scikit-learn}'s library. However, classes other than the BaseEstimator can also be used, as long as they implement the methods \textit{fit()}, \textit{predict()} and \textit{predict\_proba()}, thus enabling users to code their own local classifiers using third-party libraries, for example, \texttt{TensorFlow} \citep{abadi2016tensorflow} or \texttt{PyTorch} \citep{paszke2019pytorch}.

To allow a better integration with \texttt{scikit-learn} and faster training, \texttt{HiClass} uses for both features and labels \texttt{NumPy}'s structured arrays, whose data type is a composition of simpler data types \citep{harris2020array}. While features are exactly the same shape as expected for training flat models in \texttt{scikit-learn}, hierarchical training labels are represented as an array of shape \textit{n\_samples} $\times$ \textit{n\_levels}, where each column must contain either a label for the respective level in the hierarchy or an empty string that indicates missing labels in the leaf nodes.

The directed acyclic graph (DAG) implemented in the library \texttt{NetworkX} \citep{hagberg2008exploring} was the data structure chosen to make the models as generic as possible, since hierarchical data in the form of trees can also be represented with a DAG. Training can be performed in parallel for each local classifier by leveraging either the library \texttt{Ray}\footnotemark\footnotetext{\url{https://www.ray.io/}} or \texttt{Joblib}\footnotemark\footnotetext{\url{https://joblib.readthedocs.io/en/latest/parallel.html}}, while prediction, which is not a bottleneck, is performed from top to bottom, following the hierarchical structure to keep consistency among the several hierarchical levels.

According to \citet{silla2011survey}, the use of flat classification metrics might not be adequate to give enough insight on which algorithm is better at classifying hierarchical data. Hence, in \texttt{HiClass} we implemented the metrics of hierarchical precision (hP), hierarchical recall (hR) and hierarchical F-score (hF), which are extensions of the renowned metrics of precision, recall and F-score, but tailored to the hierarchical classification scenario. These hierarchical counterparts were initially proposed by \citet{kiritchenko2006learning}, and are defined as follows:

\vspace{0.2cm}

\centerline{
$\displaystyle{hP = \frac{\sum_i|\alpha_i\cap\beta_i|}{\sum_i|\alpha_i|}}$, $\displaystyle{hR = \frac{\sum_i|\alpha_i\cap\beta_i|}{\sum_i|\beta_i|}}$, $\displaystyle{hF = \frac{2 \times hP \times hR}{hP + hR}}$
}

\noindent , where $\alpha_i$ is the set consisting of the most specific classes predicted for test example $i$ and all their ancestor classes, while $\beta_i$ is the set containing the true most specific classes of test example $i$ and all their ancestors, with summations computed over all test examples.

\section{Code Quality Assurance}

To ensure high code quality, \textit{inheritance} is applied whenever possible to keep the code concise and easier to maintain, and all implementations adhere to the \textit{PEP 8} code style \citep{van2001pep} enforced by \texttt{flake8} and the uncompromising code formatter \texttt{black}. API documentation is provided through \textit{docstrings} \citep{goodger2010docstring}, and the implementations are accompanied by \textit{unit tests} that cover \textit{98\%} of our code and are automatically executed by our \textit{continuous integration workflow} upon commits. 

\section{Installation and Usage}

\texttt{HiClass} is hosted on GitHub\footnotemark\footnotetext{\url{https://github.com/scikit-learn-contrib/hiclass}}, while tutorials and API documentation are available on the platform Read the Docs\footnotemark\footnotetext{\url{https://hiclass.readthedocs.io/}}. Packages for Python 3.7-3.9 are available for Linux, macOS and Windows and can be obtained with \code{pip install hiclass} or \code{conda install -c conda-forge hiclass}. Code \ref{code} shows a basic example of fitting and evaluating a local hierarchical model with \texttt{HiClass}. More elaborate examples can be found in the tutorials.

\begin{lstlisting}[language=Python, label=code, caption=Example on how to use \texttt{HiClass} to train and evaluate a hierarchical classifier.]
from hiclass import LocalClassifierPerNode
from hiclass.metrics import f1
from sklearn.ensemble import RandomForestClassifier

# define mock data
X_train = X_test = [[1, 2], [3, 4]]
Y_train = Y_test = [
    ["Animal", "Mammal", "Cat"],
    ["Animal", "Reptile", "Turtle"],
]

# Use random forest classifiers for every node
rf = RandomForestClassifier()
lcpn = LocalClassifierPerNode(local_classifier=rf)
lcpn.fit(X_train, Y_train)  # Train model
predictions = lcpn.predict(X_test)  # Predict test data

# Print hierarchical F-score
print(f"f1: {f1(y_true=Y_test, y_pred=predictions)}")
\end{lstlisting}

\section{Comparison with Flat Classifiers}
\label{comparison}

\begin{figure}[!htb]
\centerline{\includegraphics[width=0.95\linewidth]{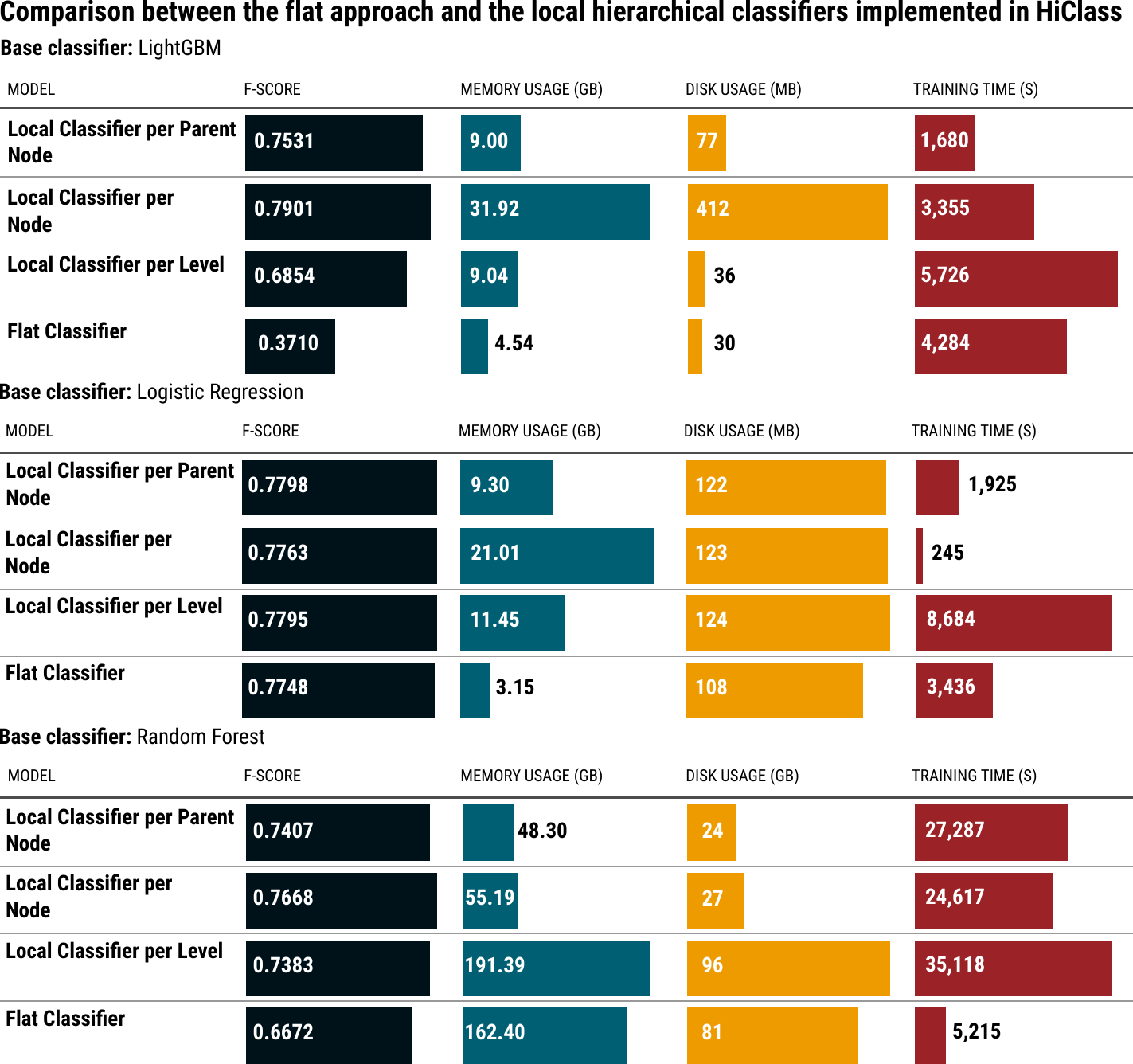}}
\caption{Comparison between the flat approach and the local hierarchical classifiers implemented in \texttt{HiClass}, using the consumer complaints data set and Microsoft's LightGBM \citep{ke2017lightgbm}, Logistic Regression and Random Forest as the base classifiers. For this benchmark, the metrics used were hierarchical F-score, memory usage in gigabyte, disk usage in megabyte or gigabyte, and training time in seconds. Overall, the hierarchical classifiers improved the F-score when compared with the flat approaches, while in some occasions the local hierarchical classifiers further reduced memory consumption, disk usage, and training time.}\label{fig:results}
\end{figure}

While \texttt{HiClass} focuses on ease of use and is fully written in a high level language, care has
been taken to maximize computational efficiency. In \autoref{fig:results}, we compare the hierarchical F-score, computational resources (measured with the command \textit{time}) and disk usage. This comparison was performed between two flat classifiers from the library \texttt{scikit-learn} and Microsoft's LightGBM \citep{ke2017lightgbm} versus the local hierarchical classifiers implemented in \texttt{HiClass}. In order to avoid bias, cross-validation and hyperparameter tuning were performed on the local hierarchical classifiers and flat classifiers. For comparison purposes, we used a snapshot from 02/11/2022 of the consumer complaints data set provided by the Consumer Financial Protection Bureau of the United States \citep{bureau2022consumer}, which after preprocessing contained 727,495 instances for cross-validation and hyperparameter tuning as well as training and 311,784 more for validation. Additional descriptions about this data set, preprocessing, feature extraction, and experimental setup are available in Appendices A-B. The benchmark was computed on multiple cluster nodes running GNU/Linux with 512 GB physical memory and 128 cores provided by two AMD EPYC\textsuperscript{\texttrademark} 7742 processors. A reproducible Snakemake pipeline \citep{koster2012snakemake} is available in our public repository\footnotemark\footnotetext{\url{https://github.com/scikit-learn-contrib/hiclass/tree/main/benchmarks/consumer_complaints}}.

Our results reveal that the hierarchical F-score was enhanced by all local hierarchical approaches when compared with the flat classifiers, achieving a maximum improvement of $\approx113\%$ when comparing the local classifier per node (LCPN) with the flat LightGBM approach. Regarding training time, the LCPN and local classifier per parent node (LCPPN) decreased computational time by $\approx93\%$ and $\approx44\%$, respectively, when compared with the flat Logistic Regression classifier. When compared with the flat Random Forest, the LCPN and LCPPN reduced memory usage by $\approx66\%$ and $\approx70\%$ and disk usage by $\approx67\%$ and $\approx70\%$, respectively.

\section{Conclusion}

 \texttt{HiClass} is a Python package that provides implementations of popular machine learning models and evaluation metrics for local hierarchical classification. Thanks to its compatibility with the \texttt{scikit-learn} API, users can choose among existing classifiers to create powerful hierarchical classifiers for hierarchical data.

In future releases, we plan to implement multi-label hierarchical classification to expand the range of problems the library can solve. Additionally, we also plan to implement the global approach, which trains a single model that learns the entire hierarchical structure. Lastly, we also intend to add support for incremental learning in order to enable training of larger than memory data sets.

\acks{The authors gratefully acknowledge all users for requesting useful new features and reporting bugs. BYR gratefully acknowledges support by the BMBF-funded Computational Life Science initiative (project DeepPath, 031L0208, to B.Y.R.).}

\appendix
\section*{Appendix A. Hierarchical Data}
\label{app:hierarchical-data}

In this appendix we delineate what constitutes hierarchical data and the data set used for evaluation in Section \ref{comparison}.

\begin{figure}[!htb]
	\centering
	\includegraphics[width=0.73\textwidth]{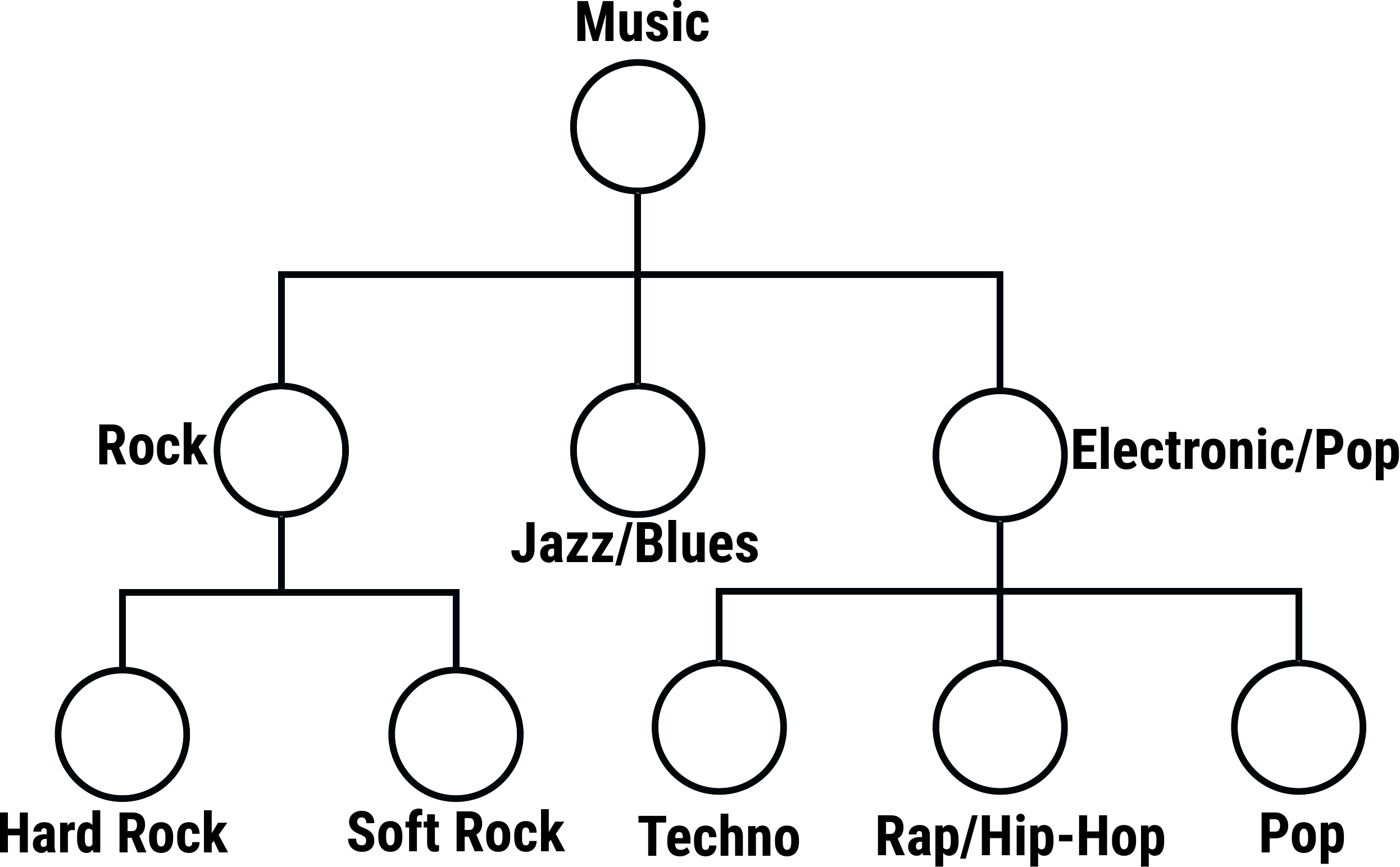}
	\caption{When labeling and retrieving musical information, the genre plays an important role, since having the musical genres structured in a
		class hierarchy simplifies how users browse and retrieve this information \citep{silla2011survey}. Image adapted from \citet{silla2011survey}.}
	\label{fig:music}
\end{figure}

\begin{figure}[!htb]
	\centering
	\includegraphics[width=0.73\textwidth]{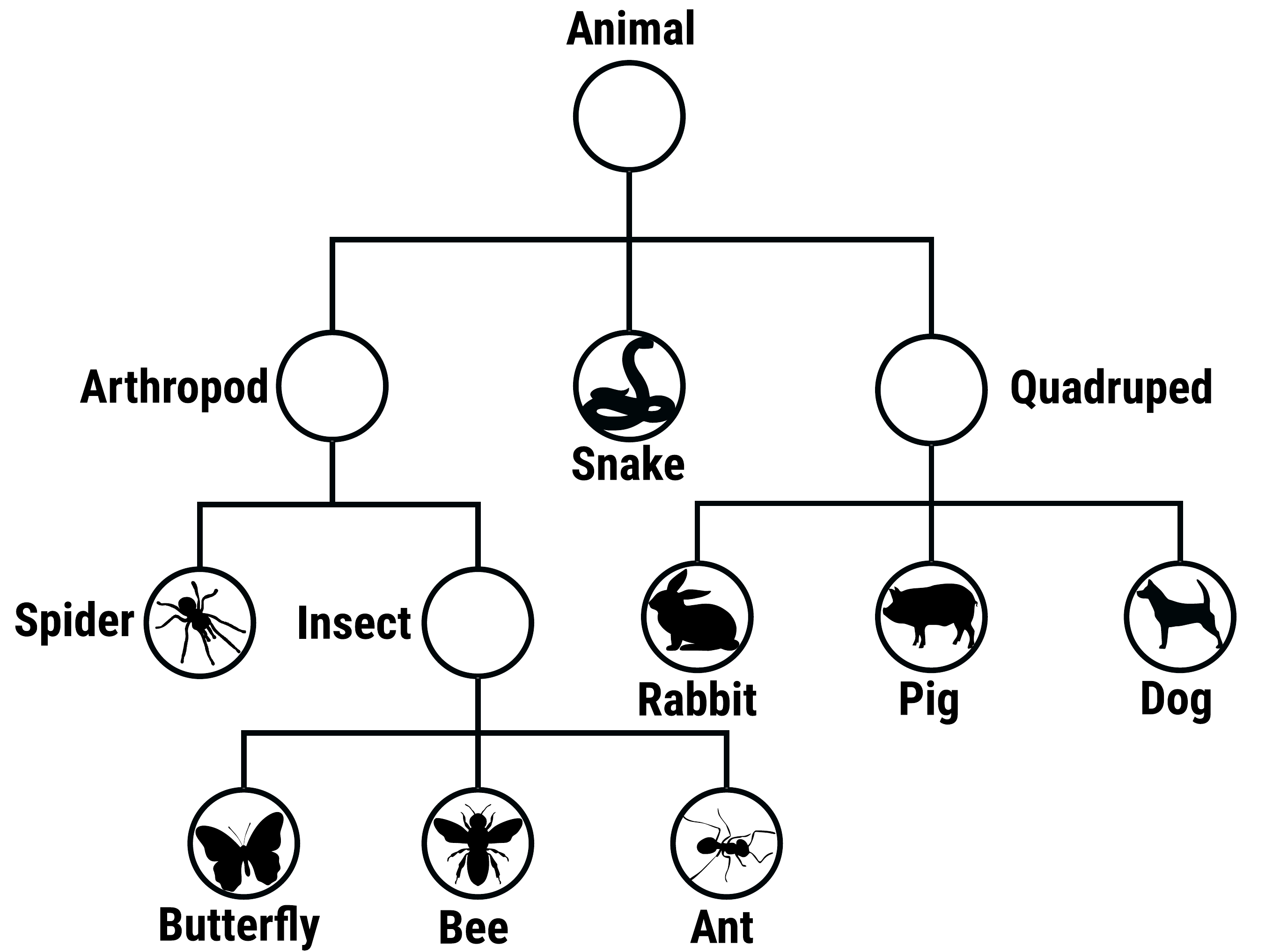}
	\caption{When grouping different species, scientists typically build a hierarchical structure to distinguish organisms according to evolutionary traits. Image inspired by \citet{barutcuoglu2006hierarchical}.}
	\label{fig:phylogeny}
\end{figure}

Numerous real-life problems can be naturally modeled hierarchically, i.e., they can be computationally represented as directed acyclic graphs or trees. Two notorious examples of hierarchical data are music genre and phylogeny, which are depicted in Figures \ref{fig:music}-\ref{fig:phylogeny}, respectively.

HiClass facilitates training local hierarchical classifiers. Hierarchical labels can be simply defined in a $m \times n$ matrix, where each row is a training example and each column is a level in the hierarchy. Such matrix can be represented with Python lists, NumPy arrays or Pandas DataFrames. Python lists and Pandas DataFrames are automatically converted to NumPy arrays for efficient processing. Training features need to be numerical, hence feature extraction might be necessary depending on the data.

\newpage

\subsection*{Consumer Complaints Data Set}
\label{app:dataset}

The consumer complaints data set is a database of complaints sent to companies in the United States \citep{bureau2022consumer}. This database, which is depicted on \autoref{tab:complaints}, is maintained by the Consumer Financial Protection Bureau (CFPB) and is updated regularly. For full reproducibility of the results, we took a snapshot from 02/11/2022 that we used for evaluation in our benchmark, and it is available upon request.


\begin{table}[!htb]
	\centering
	\begin{tabular}{@{}lll@{}}
		\toprule
		\textbf{Consumer Complaint Narrative} & \textbf{Product} & \textbf{Sub-product} \\ \midrule
		Loan & Student loan & Struggling to repay loan \\
		Credit reporting & Reports & Unable to get annual report \\ \bottomrule
	\end{tabular}
 	\caption{Depiction of consumer complaint data set, obtained from the Consumer Financial Protection Bureau \citep{bureau2022consumer}.}
	\label{tab:complaints}
\end{table}

According to the CFPB, complaints can give us insights into problems customers are experiencing and help regulate products and services, enforce laws, and educate and empower consumers. For our benchmark we used the complaint narratives as feature, which according to the CFPB are consumers' descriptions of their experiences in their own words. The classification task here consists of labeling which product and sub-product the consumer is complaining about. Classifying missing labels is a recurrent problem in various application domains, since it is common for users not to use appropriate keywords or leave fields entirely empty. For instance, automatic classification of IT tickets has become quite popular in the last couple of years \citep{revina2020ticket}, but unfortunately we could not find any comprehensive data set publicly available for a benchmark.

Feature extraction was performed using the CountVectorizer and TfidfTransformer implementations available on scikit-learn \citep{pedregosa2011scikit}, in order to compute a matrix of token counts and term-frequency, respectively, with default parameters used for both. Rows with empty cells were discarded, and around 70\% of the remaining data was used for hyperparameter tuning and training (727,495 examples), while 30\% was held for validation (311,784 examples). For the flat classifiers, labels were concatenated before training and split back into their original shape after prediction.

\section*{Appendix B. Evaluation}
\label{app:evaluation}

This appendix provides additional description about the experimental setup mentioned in Section \ref{comparison}.

\begin{figure}[!htb]
	\centering
	\includegraphics[width=0.7\textwidth]{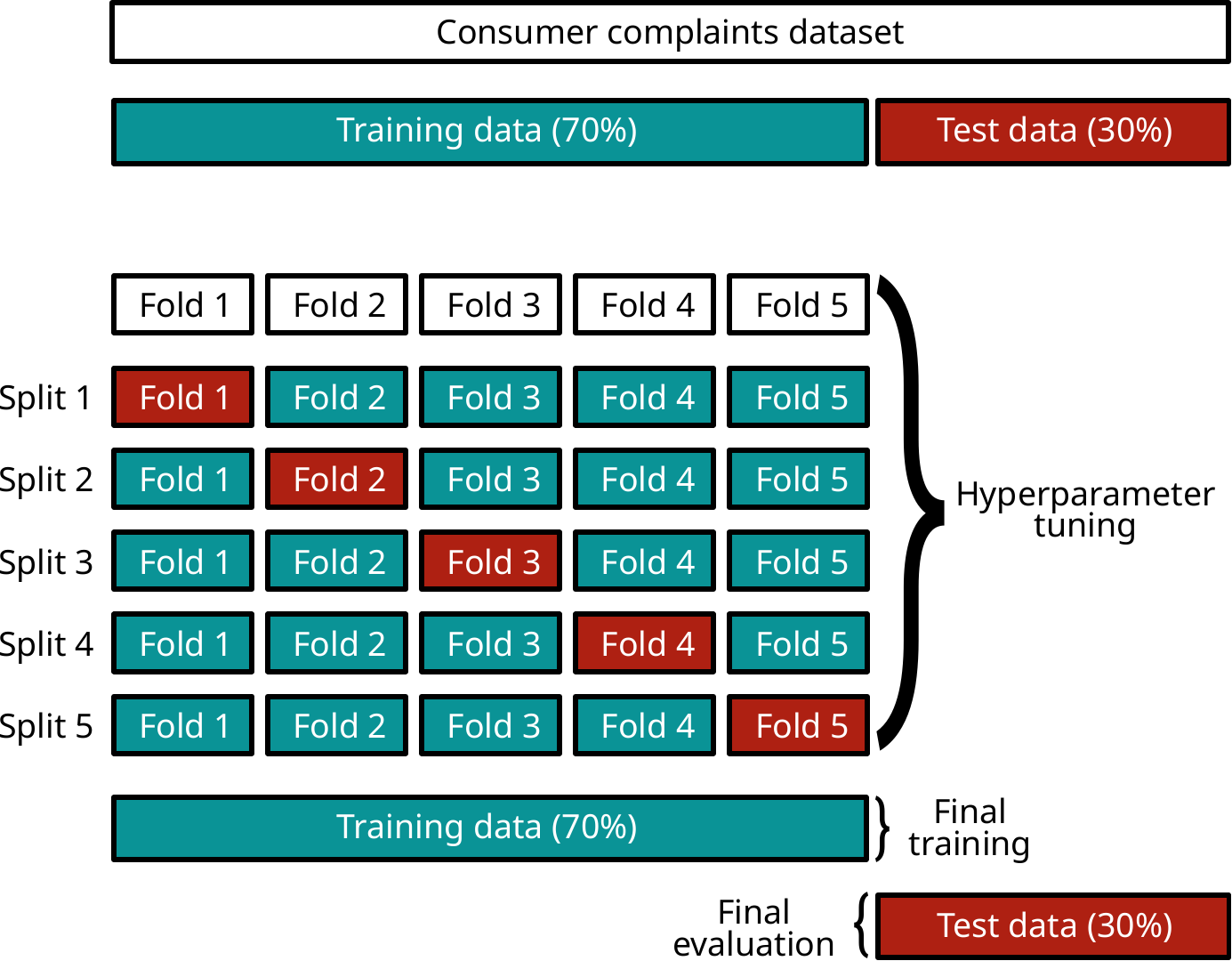}
	\caption{Visual representation of the experiments performed for hyperparameter tuning, training and evaluation. First the data set was split with 70\% of the data being used for hyperparameter tuning and training, while 30\% was held for a final evaluation. The subset with 70\% of data held for training was further split into 5 subsets for 5-fold cross-validation and identification of best hyperparameter combination. Lastly, the models were trained on the whole 70\% training data, using the best hyperparameters detected, and a final evaluation was performed with the 30\% test data held in the beginning. Image adapted from \citet{cross_validation}.}
	\label{fig:kfold}
\end{figure}

In order to avoid overfitting, our experiment began with splitting the consumer complaints data set into two subsets. The first subset contained 70\% of the data for hyperparameter tuning and training, while the remaining 30\% was held for testing (\autoref{fig:kfold}). This first division was performed via scikit-learn's train\_test\_split method. Afterwards, the training subset was further divided into 5 splits to perform 5-fold cross-validation for hyperparameter tuning. These splits were achieved with the help of the KFold class from scikit-learn.

For hyperparameter tuning, the models were trained using 4 folds as training data and validated on the remaining one. This process was repeated 5 times, with different folds combinations being used in each iteration, and the average hierarchical F-score was reported as the performance metric. The selection of the best hyperparameters was assisted by Hydra \citep{hydra} and its plugin Optuna \citep{optuna}, through a grid search using the combinations of hyperparameters described in Tables \ref{tab:lightgbm}-\ref{tab:random_forest}. After the best hyperparameter combinations were identified, the models were trained once more, but this time using the entire training data (70\% of the full data set) and a final evaluation was carried out on the test data held in the beginning (30\% of the entire data). All these steps were automatically executed in a Snakemake pipeline, which is available in our repository\footnotemark\footnotetext{\url{https://github.com/scikit-learn-contrib/hiclass/tree/main/benchmarks/consumer_complaints}}.

\begin{table}[!htb]
\centering
\begin{tabular}{@{}ccc@{}}
\toprule
\textbf{num\_leaves} & \textbf{n\_estimators} & \textbf{min\_child\_samples} \\ \midrule
31 & 100 & 20 \\
62 & 200 & 40 \\ \bottomrule
\end{tabular}
\caption{Hyperparameters tested for LightGBM.}
\label{tab:lightgbm}
\end{table}

\begin{table}[!htb]
\centering
\begin{tabular}{lc}
\hline
\textbf{solver} & \textbf{max\_iter} \\ \hline
newton-cg & 10000 \\
lbfgs &  \\
liblinear &  \\
sag &  \\
saga &  \\ \hline
\end{tabular}
\caption{Hyperparameters tested for logistic regression.}
\label{tab:logistic_regression}
\end{table}

\begin{table}[!htb]
\centering
\begin{tabular}{@{}cl@{}}
\toprule
\textbf{n\_estimators} & \textbf{criterion} \\ \midrule
100 & gini \\
200 & entropy \\
 & log\_loss \\ \bottomrule
\end{tabular}
\caption{Hyperparameters tested for random forest.}
\label{tab:random_forest}
\end{table}

\newpage

\begin{figure}[!b]
	\centering
	\includegraphics[width=\textwidth]{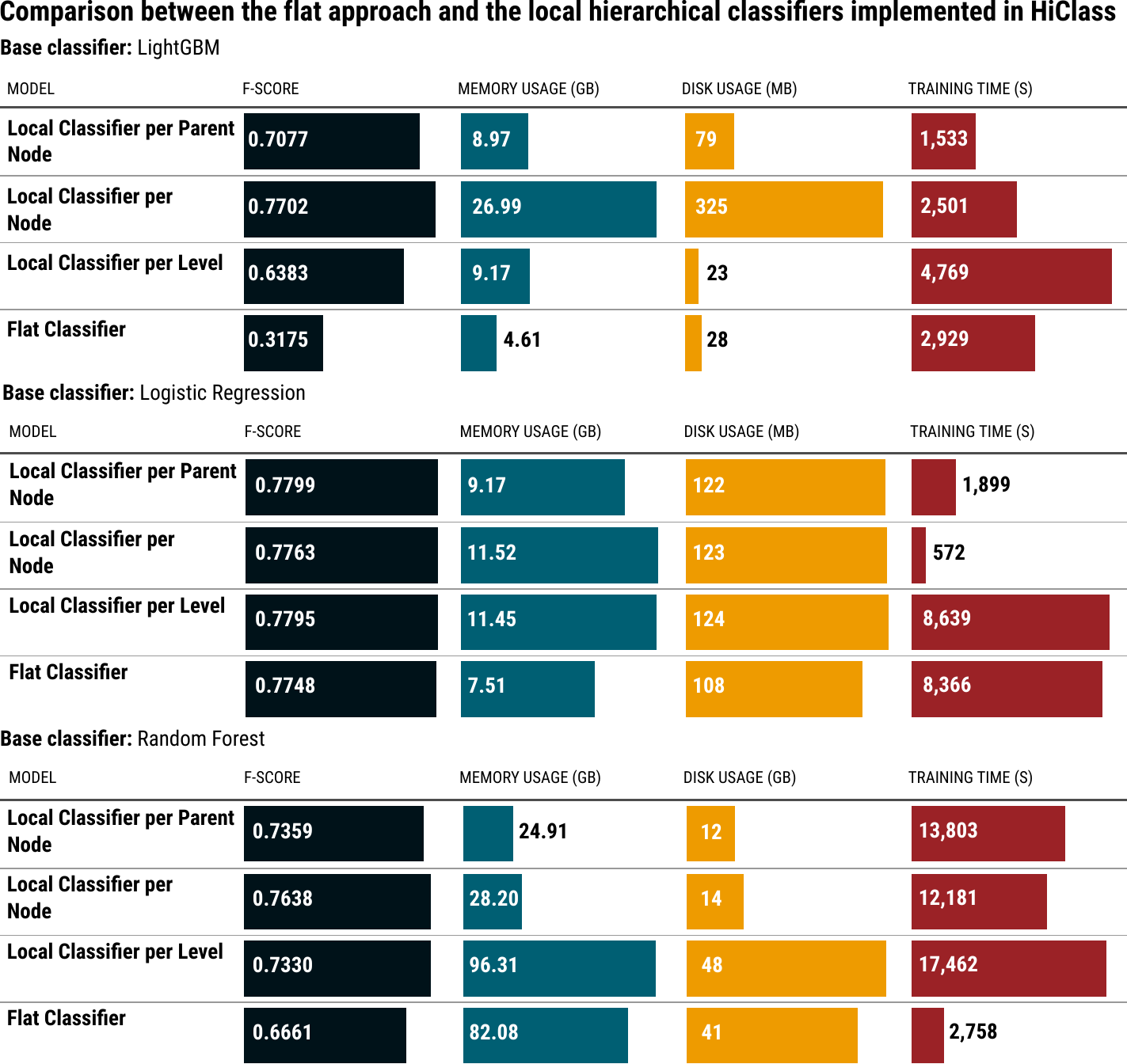}
	\caption{Comparison between the flat approach and the local hierarchical classifiers implemented in \texttt{HiClass}, using the consumer complaints data set and Microsoft's LightGBM \citep{ke2017lightgbm}, Logistic Regression and Random Forest as the base classifiers. For this benchmark, the metrics used were hierarchical F-score, memory usage in gigabyte, disk usage in megabyte or gigabyte, and training time in seconds. No hyperparameter tuning was performed for these results. Overall, the hierarchical classifiers improved the F-score when compared with the flat approach, while in some occasions the local hierarchical classifiers further reduced memory consumption, disk usage, and training time.}
	\label{fig:no_tuning}
\end{figure}

For comparison purposes, in \autoref{fig:no_tuning} we show the results without hyperparameter tuning.

\section*{Appendix C. Algorithms Overview}
\label{app:algorithms}

This appendix provides rigorous descriptions for the algorithms mentioned in earlier sections.

HiClass provides implementations for the most popular machine learning models for local hierarchical classification, including the Local Classifier Per Node, the Local Classifier Per Parent Node and the Local Classifier Per Level. In the following subsections, we present in more details these different approaches for local hierarchical classification.

\subsection*{Local Classifier Per Node}
\label{app:lcpn}
\begin{figure}[!htb]
	\centering
	\includegraphics[width=0.73\textwidth]{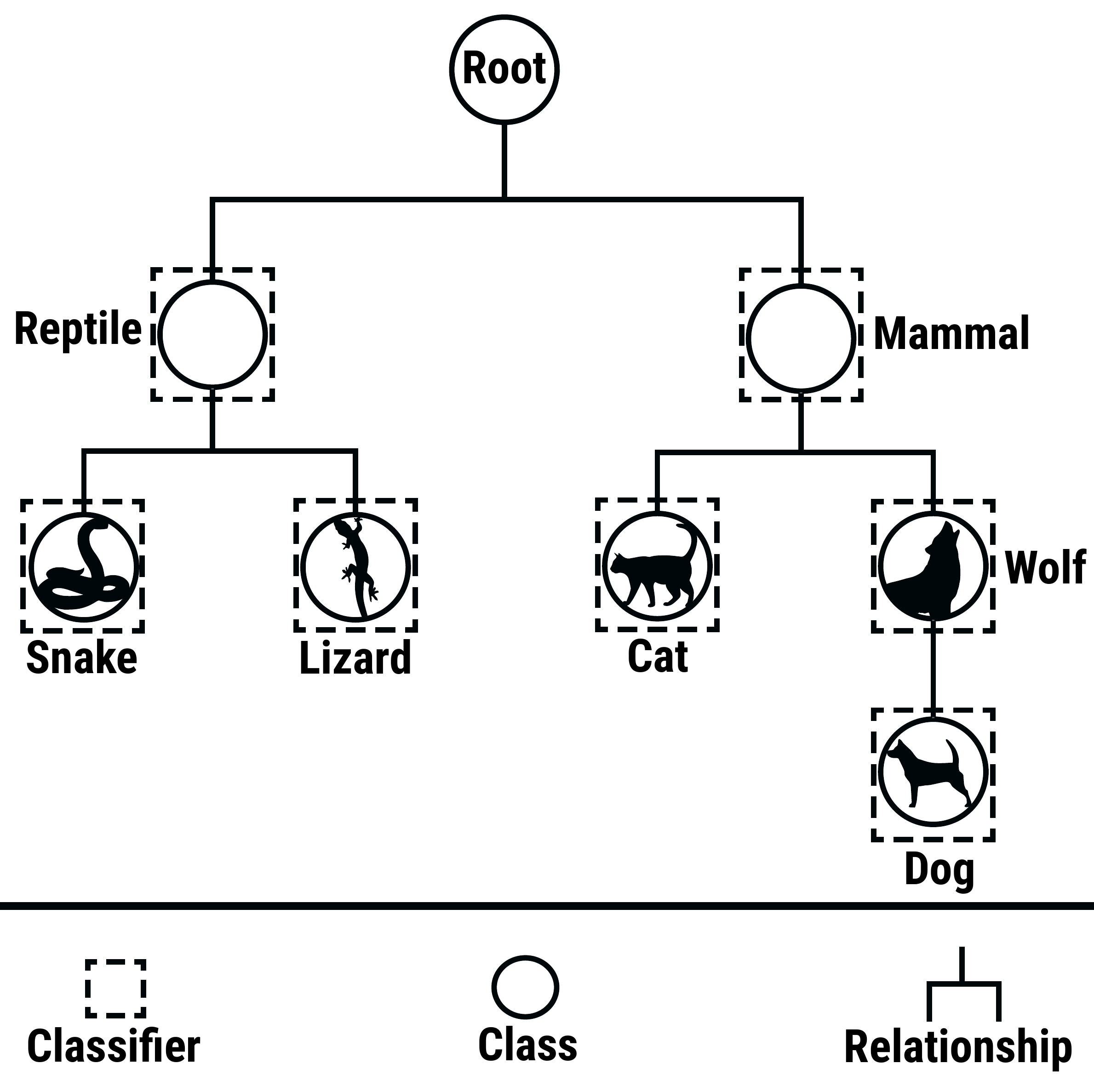}
	\caption{Visual representation of the local classifier per node approach, where binary classifiers (squares) are trained for each class (circles) of the hierarchy, excluding the root node.}
	\label{fig:lcpn}
\end{figure}

One of the most popular approaches in the literature, the local classifier per node consists of training one binary classifier for each node of the class taxonomy, except for the root node. A visual representation of the local classifier per node is shown in \autoref{fig:lcpn}.

Each binary classifier can be trained in parallel using either the library Ray or Joblib. In order to avoid inconsistencies, prediction is performed in a top-down manner. For example, given a hypothetical test example, the local classifier per node firstly queries the binary classifiers at nodes “Reptile” and “Mammal”. Supposing that in this hypothetical situation the probability of the test example belonging to class “Reptile” is 0.8, and the probability of belonging to class “Mammal” is 0.5, then class “Reptile” is selected for the first level. At the next level, only the classifiers at nodes “Snake” and “Lizard” are queried, and again the one with the highest probability is chosen.

\newpage

\subsubsection*{Training Policies}
\label{app:policies}

There are multiple ways to define the set of positive and negative examples for training the binary classifiers. In HiClass we implemented 6 policies described by \citet{silla2011survey}, which were based on previous work from \citet{eisner2005improving} and \citet{fagni2007selection}. The notation used to define the sets of positive and negative examples is presented in \autoref{tab:notation}, as described by \citet{silla2011survey}.

\begin{table}[!htb]
	\centering
	\begin{tabular}{@{}ll@{}}
		\toprule
		\textbf{Symbol} & \textbf{Meaning} \\ \midrule
		$Tr$ & The set of all training examples \\
		$Tr^+(c_i)$ & The set of positive training examples of $c_i$ \\
		$Tr^-(c_i)$ & The set of negative training examples of $c_i$ \\
		$\uparrow (c_i)$ & The parent category of $c_i$ \\
		$\downarrow (c_i)$ & The set of children categories of $c_i$ \\
		$\Uparrow (c_i)$ & The set of ancestor categories of $c_i$ \\
		$\Downarrow (c_i)$ & The set of descendant categories of $c_i$ \\
		$\leftrightarrow (c_i)$ & The set of sibling categories of $c_i$ \\
		$*(c_i)$ & Denotes examples whose most specific known class is $c_i$ \\
		\bottomrule
	\end{tabular}
	\caption{Notation used to define the sets of positive and negative examples.}
	\label{tab:notation}
\end{table}

\newpage

Based on this notation, we can define the different policies and their sets of positive and negative examples as follows:

\begin{table}[!htb]
	\centering
	\begin{tabular}{@{}lll@{}}
		\toprule
		\textbf{Policy} & \textbf{Positive examples} & \textbf{Negative examples} \\ \midrule
		\textbf{Exclusive} & $Tr^+(c_i) = *(c_i)$ & $Tr^-(c_i) = Tr \setminus *(c_i)$ \\
		\textbf{Less exclusive} & $Tr^+(c_i) = *(c_i)$ & $Tr^-(c_i) = Tr \setminus *(c_i) \cup \Downarrow (c_i)$ \\
		\textbf{Less inclusive} & $Tr^+(c_i) = *(c_i) \cup \Downarrow (c_i)$ & $Tr^-(c_i) = Tr \setminus *(c_i) \cup \Downarrow (c_i)$ \\
		\textbf{Inclusive} & $Tr^+(c_i) = *(c_i) \cup \Downarrow (c_i)$ & $Tr^-(c_i) = Tr \setminus *(c_i) \cup \Downarrow (c_i) \cup \Uparrow (c_i)$ \\
		\textbf{Siblings} & $Tr^+(c_i) = *(c_i) \cup \Downarrow (c_i)$ & $Tr^-(c_i) = \leftrightarrow (c_i) \cup \Downarrow (\leftrightarrow (c_i))$ \\
		\textbf{Exclusive siblings} & $Tr^+(c_i) = *(c_i)$ & $Tr^-(c_i) = \leftrightarrow (c_i)$ \\
		\bottomrule
	\end{tabular}
	\caption{Policies used to define the sets of positive and negative examples.}
	\label{tab:policies}
\end{table}

Using as example the class “Wolf” from the hierarchy represented in \autoref{fig:lcpn}, we have the following sets of positive and negative examples for each policy:

\begin{table}[!htb]
	\centering
	\begin{tabular}{@{}lll@{}}
		\toprule
		\textbf{Policy} & $Tr^+(c_{\text{Wolf}})$ & $Tr^-(c_{\text{Wolf}})$ \\ \midrule
		\textbf{Exclusive} & Wolf & Reptile, Snake, Lizard, Mammal, Cat, Dog \\
		\textbf{Less exclusive} & Wolf & Reptile, Snake, Lizard, Mammal, Cat \\
		\textbf{Less inclusive} & Wolf, Dog & Reptile, Snake, Lizard, Mammal, Cat \\
		\textbf{Inclusive} & Wolf, Dog & Reptile, Snake, Lizard, Cat \\
		\textbf{Siblings} & Wolf, Dog & Cat \\
		\textbf{Exclusive siblings} & Wolf & Cat \\
		\bottomrule
	\end{tabular}
	\caption{Sets of positive and negative examples for each policy, given the class "Wolf".}
	\label{tab:example_wolf}
\end{table}

\subsection*{Local Classifier Per Parent Node}
\label{app:lcppn}

\begin{figure}[!htb]
	\centering
	\includegraphics[width=0.73\textwidth]{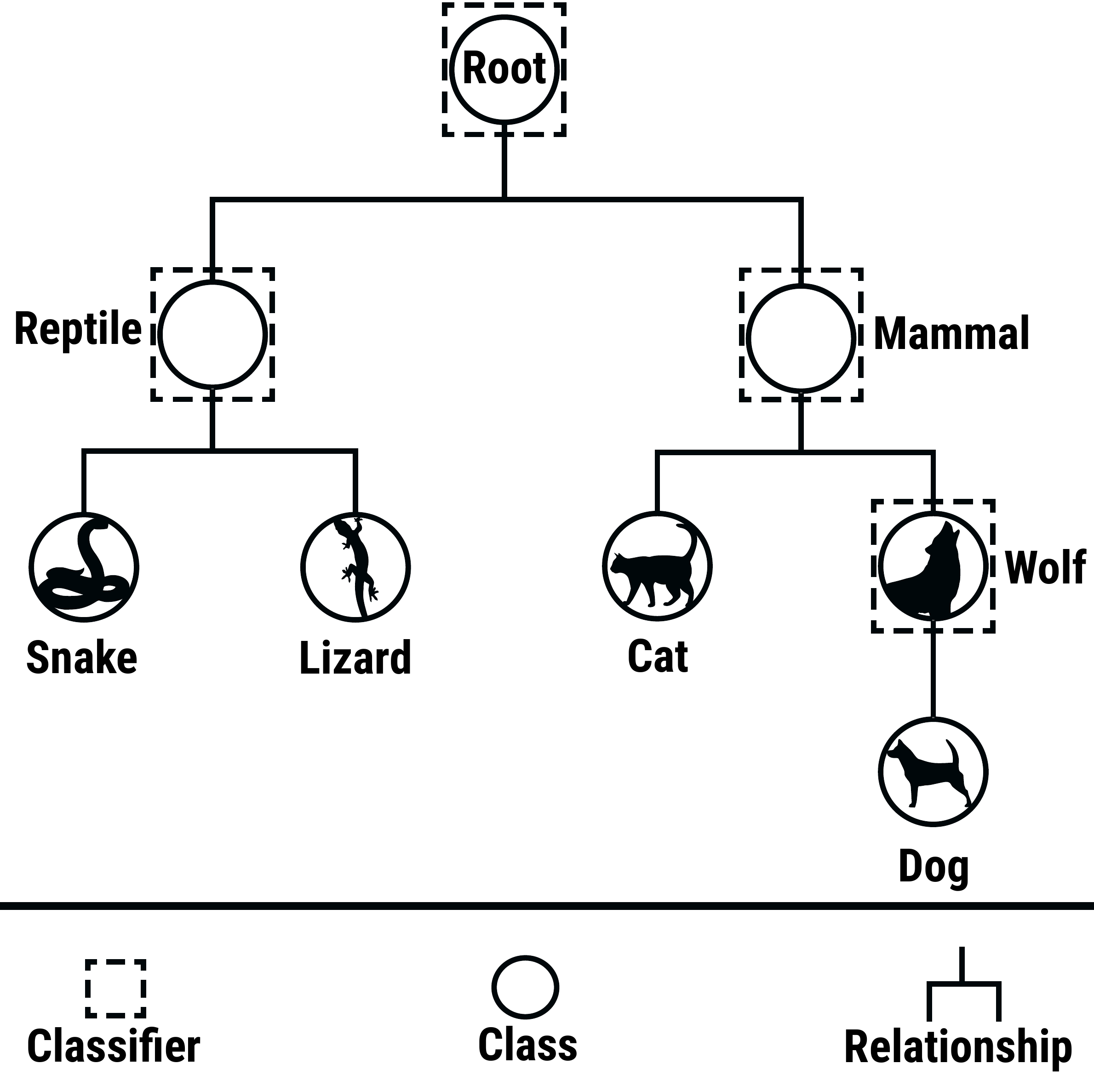}
	\caption{Visual representation of the local classifier per parent node approach, where multi-class classifiers (squares) are trained for each parent node existing in the class hierarchy (circles).}
	\label{fig:lcppn}
\end{figure}

The local classifier per parent node approach consists of training a multi-class classifier for each parent node existing in the hierarchy, as shown in \autoref{fig:lcppn}.

While training can be executed in parallel, prediction is always performed in a top-down style in order to avoid inconsistencies. For example, assuming that the classifier located at the root node decides that a test example belongs to class “Mammal”, then the next level can only be predicted by the classifier located at node “Mammal”, which in turn will ultimately decide if the test example belongs either to the class “Cat” or “Wolf/Dog”.

\subsection*{Local Classifier Per Level}
\label{app:lcpl}

\begin{figure}[!htb]
	\centering
	\includegraphics[width=0.73\textwidth]{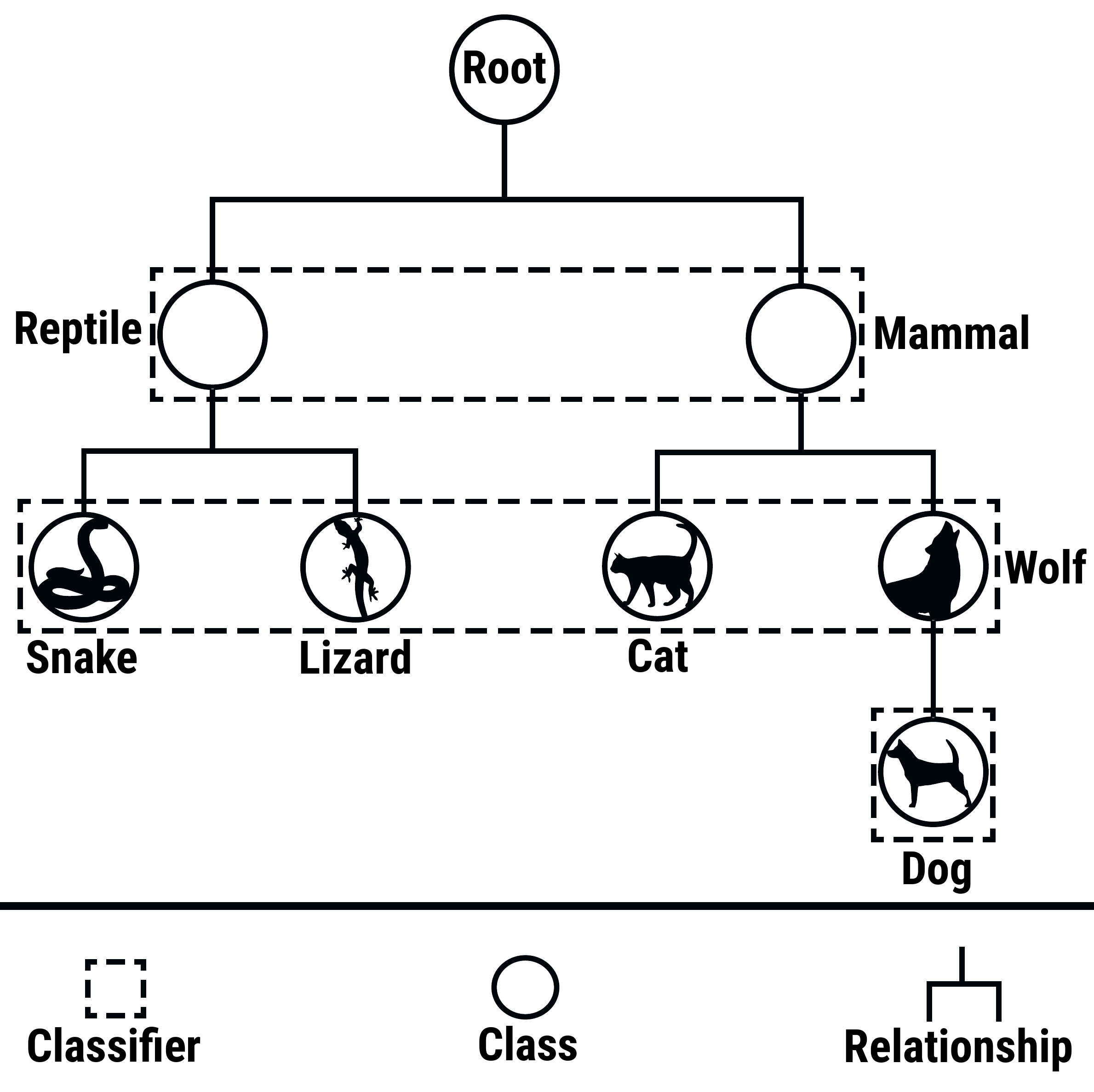}
	\caption{Visual representation of the local classifier per level approach.}
	\label{fig:lcpl}
\end{figure}

The local classifier per level approach consists of training a multi-class classifier for each level of the class taxonomy. An example is displayed on \autoref{fig:lcpl}.

Similar to the other hierarchical classifiers, the local classifier per level can also be trained in parallel, and prediction is performed in a top-down mode to avoid inconsistencies. For example, supposing that for a given test example the classifier at the first level returns the probabilities 0.91 and 0.7 for classes “Reptile” and “Mammal”, respectively, then the one with the highest probability is considered as the correct prediction, which in this case is class “Reptile”. For the second level, only the probabilities for classes “Snake” and “Lizard” are considered and the one with the highest probability is the final prediction.

\vskip 0.2in
\bibliography{manuscript}

\end{document}